%% file: paper.tex
\documentclass{article} 
\usepackage{iclr2024_conference,times}
\usepackage{CJKutf8}
\usepackage{rotating}
\usepackage{multirow}
\usepackage{wrapfig}
\usepackage{lipsum}
\usepackage{makecell}
\input{math_commands.tex}

\input{packages.tex}

\title{From Beginner to Expert: Modeling Medical Knowledge into General LLMs}

\author{\makecell[l]{Qiang Li\textsuperscript{$\dagger$}, Xiaoyan Yang\textsuperscript{$\dagger$}, Haowen Wang, Qin Wang, Junjie Wang, Yang Zhang,\\Mingyuan Chu, Sen Hu, Yicheng Chen, Yue Shen, Cong Fan, Wangshu Zhang,\\Teng Xu, Jinjie Gu, Jing Zheng, Guannan Zhang}\\
Ant Group\\
\texttt{\{mangxiao.lq,joyce.yxy,zhanying\}@antgroup.com}\\
\AND
Lei Liu\textsuperscript{$\ddagger$,$\bigtriangleup$} \\
Ant Group, The Chinese University of Hong Kong, Shenzhen (CUHK-SZ) \\
\texttt{\{liulei1497\}@gmail.com}\\
}

\iclrfinalcopy 
\begin{document}
\maketitle

\newcommand \footnoteONLYtext[1]
{
	\let \mybackup \thefootnote
	\let \thefootnote \relax
	\footnotetext{#1}
	\let \thefootnote \mybackup
	\let \mybackup \imareallyundefinedcommand
}
\footnoteONLYtext{\textsuperscript{$\dagger$}These authors contributed equally to this work.} 
\footnoteONLYtext{\textsuperscript{$\ddagger$}Corresponding author.}
\footnoteONLYtext{\textsuperscript{$\bigtriangleup$}Work was done during Lei Liu's research internship in Ant Group.}

\begin{abstract}

Recently, large language model (LLM) based artificial intelligence (AI) systems have demonstrated remarkable capabilities in natural language understanding and generation. However, these models face a significant challenge when it comes to sensitive applications, such as reasoning over medical knowledge and answering medical questions in a physician-like manner. Prior studies attempted to overcome this challenge by increasing the model size (>100B) to learn more general medical knowledge, while there is still room for improvement in LLMs with smaller-scale model sizes (<100B). In this work, we start from a pre-trained general LLM model (AntGLM-10B) and fine-tune it from a medical beginner towards a medical expert (called AntGLM-Med-10B), which leverages a 3-stage optimization procedure, \textit{i.e.}, general medical knowledge injection, medical domain instruction tuning, and specific medical task adaptation. Our contributions are threefold: (1) We specifically investigate how to adapt a pre-trained general LLM in medical domain, especially for a specific medical task. (2) We collect and construct large-scale medical datasets for each stage of the optimization process. These datasets encompass various data types and tasks, such as question-answering, medical reasoning, multi-choice questions, and medical conversations. (3) Specifically for multi-choice questions in the medical domain, we propose a novel Verification-of-Choice approach for prompting engineering, which significantly enhances the reasoning ability of LLMs. Remarkably, by combining the above approaches, our AntGLM-Med-10B model can outperform the most of LLMs on PubMedQA, including both general and medical LLMs, even when these LLMs have larger model size.

\end{abstract}

\section{Introduction}

The ability of large language models (LLMs) is truly remarkable to understand and generate text in various fields like natural language, computer code, and protein sequences. These LLMs leverage the transformer architecture~\citep{vaswani2017attention}, which is specifically designed for sequence modeling and trained through self-supervision~\citep{kenton2019bert}. By increasing the model size, dataset size, and training computation, the performance on various benchmarks are consistently improved~\citep{liang2022holistic}. These empirical findings are in line with a theoretical analysis~\citep{kaplan2020scaling}, which highlights the significance of scale in ensuring the reliability of inferences made by LLMs.

It is a long-standing research topic for AI in medicine to develop LLMs for solving medical problems, where accurate assessment of medical knowledge and reasoning capabilities is crucial for informed decision-making and favorable patient outcomes. Currently, LLMs for applications in medicine usually fail to fully utilize medical domain data, due to lacking general and specific clinical knowledge~\citep{yim2020predicting}. As indicated by~\citep{singhal2022large}, there is a discordance between what AI models can do and what may be expected of them in real-world clinical workflows~\citep{lakkaraju2022rethinking, schaekermann2020expert}. 

There are two kinds of exploration paths to investigate the adaptation of LLMs in the medical domain. One approach is to conduct a thorough assessment of general language models like GPT-3.5~\citep{brown2020language}, GPT-4~\citep{openai2023gpt4}, and ChatGPT~\citep{chatgpt}, without any specific fine-tuning for medical clinical issues. These models are designed for general purposes and are not specialized for medical domain. In~\citep{nori2023capabilities}, researchers evaluated performance of GPT-4 model with its predecessors in the GPT family on medical problems. Another way is to fine-tune a specialized LLM model through training or engineered to solve medical clinical tasks. For example, \cite{singhal2023expertlevel} developed a new medical LLM called Med-PaLM 2 and targeted medical domain-specific fine-tuning, which is based on a new base model (PaLM 2 \cite{anil2023palm}). To evaluate how effectively LLMs encodes clinical medical knowledge, previous works~\citep{jin-etal-2019-pubmedqa,singhal2023expertlevel} generally considered the medical question answering task, which requires deep understanding on medical context, accurately recalling relevant medical knowledge, and reasoning with expert-level experiences.


Nevertheless, existing medical LLMs are mainly based on scaling law~\citep{chung2022scaling} to train a larger model over massive data, which indeed lacks a fundamental optimization paradigm to adapt a pre-trained general language model towards a medical-specific expert. We conjecture that it is mainly due to the intrinsic LLM training paradigm, \textit{i.e.}, large-scale pre-training followed by specific fine-tuning. Concretely, as a medical beginner, a pre-trained general LLM needs to learn basic medical knowledge as background, which requires a continual pre-training over the large-scale medical data. Then, considering diverse task types on medical domain (\textit{e.g.}, QA, Multi-choice Question, and Reasoning), a relatively large-scale instruction fine-tuning process should be applied to encode task-related knowledge into LLMs. Finally, given a medical problem with a specific task type, a careful fine-tuning step can help to quickly and accurately adjust a LLM as a medical expert.

\begin{figure}[!t]
    \centering
    \includegraphics[width=1\linewidth]{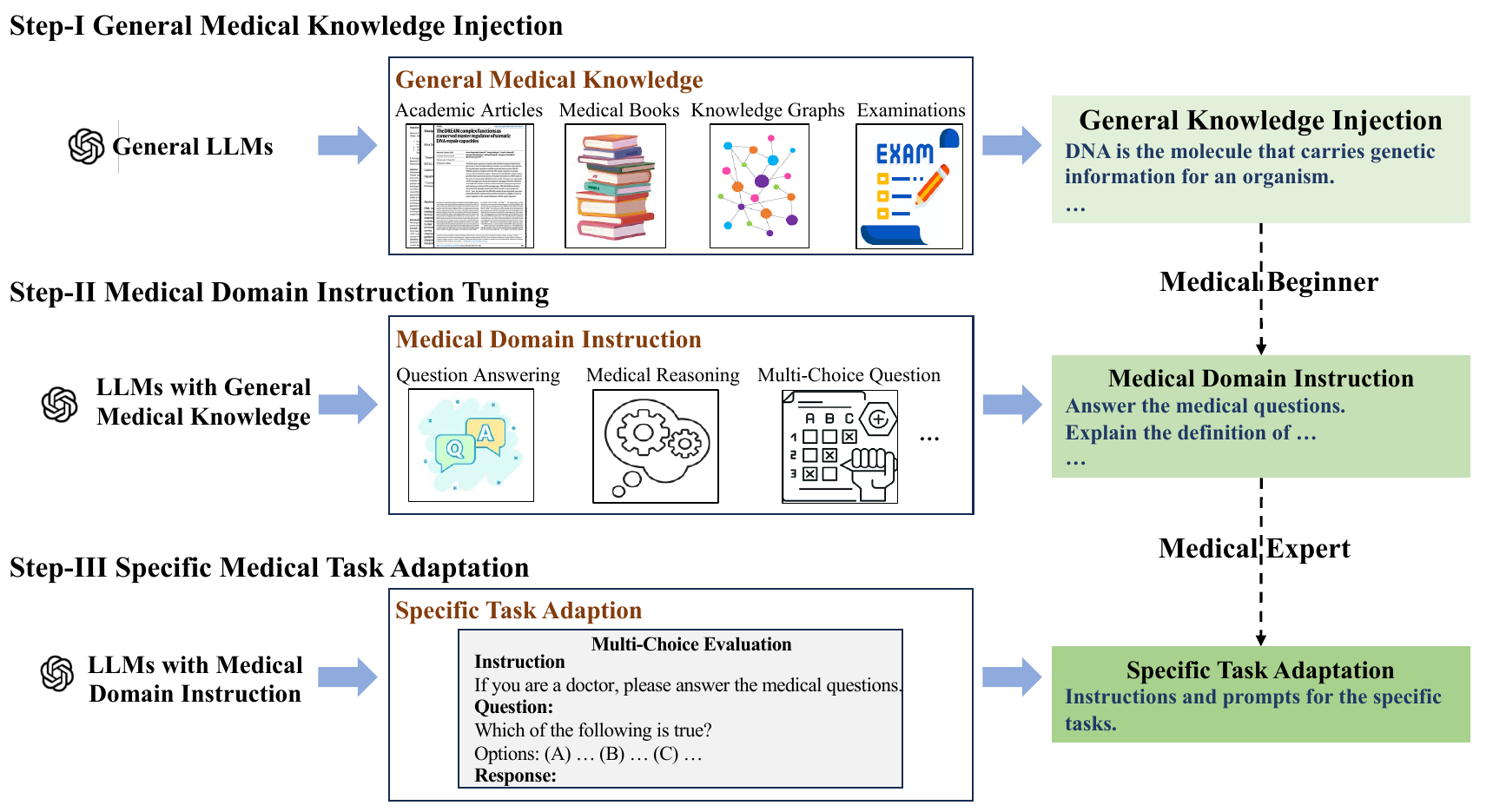}
    \caption{The 3-stage optimization procedure of AntGLM-Med-10B. We collect different data types (\textit{e.g.}, medical articles, books, and examinations) and medical tasks (question answering, reasoning, and multi-choice questions). In the adaptation stage, we mainly consider the optimizations for multi-choice questions, resulting in competitive performance on the PubMedQA. }
    \label{fig:framework}
\end{figure}

In this study, we present a comprehensive methodology for fine-tuning a pre-trained general LLM model to transform it from a medical beginner to a medical expert. As shown in Figure \ref{fig:framework}, this process involves a 3-stage optimization procedure, namely continual pre-training for medical knowledge injection, medical domain instruction tuning, and specific medical task adaptation. To support each stage of fine-tuning, we curate and construct diverse large-scale medical datasets that encompass various data types and cover different tasks. These tasks include question-answering (QA), medical reasoning, multi-choice question, and medical conversations. Additionally, for the multi-choice question task within the medical domain, we introduce a novel Verification-of-Choice approach for prompting engineering. This approach significantly enhances the reasoning ability of LLMs, offering a valuable contribution to the field. By incorporating the afore-mentioned components, the obtained AntGLM-Med-10B can achieve an impressive accuracy on the PubMedQA. Notably, AntGLM-Med-10B can outperform many larger LLMs (>40B), demonstrating its effectiveness and potential in the medical domain.

The organization of this paper is as follows: 

\textbf{(1)} In Section \ref{related_work}, we provide a overview for the recent progress of LLMs and discuss some applications of LLMs in the medical domain. 

\textbf{(2)} In Section \ref{framework_formulation}, we illustrate the framework preliminaries in detail, including the 3-stage optimization (continual pre-training, instruction fine-tuning, and specific-task adaptation), dataset collection and construction, and the utilized techniques. 

\textbf{(3)} In Section \ref{experiment}, we provide comprehensive experiments to indicate the effectiveness of 3-stage optimization, as well as the well analysis for each tuning strategies. The performance on PubMedQA is significantly improved.

\newpage

\section{Related Work}
\label{related_work}
\textbf{Large Language Models.} LLMs have achieved remarkable performance in various natural language processing (NLP) tasks~\citep{brown2020language,wei2022chain,liu2023thinkinmemory}, in recent years. These developments benefits from scaling up the training of both model size (typically for transformer-based models) and data scale, where scaling law \cite{chung2022scaling} indicates the relationship between model performance and model scale and dataset size. For example, when trained on extensive text corpora like Wikipedia and BooksCorpus, LLMs could show promising performance across various NLP tasks, including tasks requiring specialized domain knowledge and reasoning. In detail, GPT-3~\citep{brown2020language} stands out as the pioneering language model with impressive 100 billion parameters, which showcases remarkable few-shot learning capabilities and introduces the concept of in-context learning. Following its success, a plethora of other LLMs have been proposed, such as Megatron-LM~\citep{korthikanti2023reducing}, OPT~\citep{zhang2022opt}, Chinchilla~\citep{hoffmann2022training}, Galactica~\citep{taylor2022galactica}, LLaMA~\citep{touvron2302llama}, PaLM~\citep{chowdhery2022palm}, and PaLM-2~\citep{singhal2023expertlevel}. These LLM models have further enhanced language understanding, generation, instruction following, reasoning abilities, and even possess a deep understanding of common sense knowledge~\citep{mao2023gpteval}. As a result, they have become indispensable base models across various domains, including Finance~\citep{yang2023fingpt}, Education~\citep{milano2023large}, and Healthcare \cite{arora2023promise}.

\textbf{Medical LLMs.} These advanced improvements for LLMs have also demonstrated the effectiveness on medical domain, such as HuatuoGPT~\citep{zhang2023huatuogpt}, Med-PaLM 2~\citep{singhal2023expertlevel}, and Visual Med-Alpaca~\citep{gao2023ophglm}. In particular, HuatuoGPT~\citep{zhang2023huatuogpt} presented to actively ask questions for the patients rather than only make a respond. Visual Med-Alpaca~\citep{gao2023ophglm} integrated visual experts with LLMs for multi-modal biomedical tasks, which can perform better over various tasks. Although these approaches utilized scientific and biomedical corpora for both discriminative and generative language modeling, they are typically small in model size compared with LLMs GPT-3~\citep{brown2020language} and PaLM~\citep{chowdhery2022palm}.

\section{From Beginner to Expert}
\label{framework_formulation}
\subsection{Framework Formulation}
In this work, our objective is to teach a medical beginner (\textit{i.e.}, a pre-trained general LLM) and let it become a medical expert (\textit{i.e.}, a fine-tuned medical LLM), which corresponds to a full procedure for adapting a pre-trained foundational large language model in the medical domain. The optimization process can be divided into three key steps: General Medical Knowledge Injection, Medical Domain Instruction Tuning, and Specific Medical Task Adaptation.

\begin{table}[!t]
\centering
\caption{The collected datasets for different optimization stages.}
\label{medqa}
\resizebox{\linewidth}{!}{
\begin{tabular}{cll}
\toprule
\textbf{Optimization}  & \textbf{Dataset} & \textbf{Description} \\ \midrule
\multirow{5}{*}{\makecell[c]{General Medical \\ Knowledge Injection}} 
& Medical Books &  General medical knowledge in medical and science textbooks \\ \cmidrule{2-3}
& Knowledge Graphs & Highly structured medical knowledge in open-source knowledge graphs \\ \cmidrule{2-3}
& Question-Answer Pairs & Real-world medical consultation information in textual form \\ \cmidrule{2-3}
& Exam Questions & Question, answer, and explanation for testing medical knowledge points \\ \cmidrule{2-3}
& Articles & Professional medical and science articles written by different doctors\\ \midrule

\multirow{4}{*}{\makecell[c]{Medical Domain \\ Instruction Tuning}} 
& PromptCBLUE &  A instruction-tuning dataset for multi-task and few-shot learning in Chinese \\ \cmidrule{2-3}
& Chinese Examination & A dataset collected from the Chinese physician examination data \\ \cmidrule{2-3}
& Wuma QA & A large-scale QA database covering encyclopedia, hospital, and doctor \\ \cmidrule{2-3}
& Huatuo Wiki &  A subset (the data source is Chinese Wikipedia) of Huatuo-26M \\ \cmidrule{2-3}
& Multiple-choice Question & Multiple-choice questions from different datasets \\ \midrule

Specific Medical Task Adaptation & Multiple-choice Question & Multiple-choice questions from different datasets \\ \bottomrule
\end{tabular}}
\end{table}

As shown in Figure \ref{fig:framework}, \textbf{General Medical Knowledge Injection} aims to encode the fundamental medical knowledge into the pre-trained language model. \textbf{Medical Domain Instruction Tuning} can enrich the language model with diverse medical task types. \textbf{Specific Medical Task Adaptation} can tailor the model to align with a specific clinical task.

\subsection{Base LLM: AntGLM}
The base LLM in this work is AntGLM, a general large-scale model developed by Ant Group. We conduct further pre-training and fine-tuning for its adaptations on the medical domain.


\textbf{Architecture.} AntGLM is based on the GLM architecture~\citep{DBLP:conf/acl/DuQLDQY022}, which combines the ideas of auto-encoding and auto-regression to enhance the learning process. AntGLM is featured by 48 transformer layers, a hidden size of 4096, and 64 attention heads, resulting in 10B parameters. AntGLM incorporates two-dimensional positional encoding and enables the pre-training task of predicting the order of blank regions, which can significantly improve the performance of blank filling during pre-training in a flexible manner. Overall, AntGLM performs well on various natural language processing tasks.

\subsection{Datasets for General Medical Knowledge Injection}
As introduced below, several authoritative Chinese and English datasets are collected for the stage of General Medical Knowledge Injection, which mainly include PromptCBLUE~\citep{zhu2023promptcblue}, MedPaLM2~\citep{singhal2023expertlevel} dataset, and 3 sets of Chinese examination datasets.



\textbf{Medical Books.} In particular, we collect some medical books as training corpora, which include medical textbooks and popular science books. The high-quality and lengthy text in the book data makes LLMs easier to learn semantic correlations among contexts. We manually remove the contents in non-text format from the books, such as tables and appendices, which may influence the model training. After data pre-processing, we used approximately 150,000 medical-related books during pre-training.

\textbf{Knowledge Graphs.} We use an open-source medical knowledge graph to generate pre-training data, \textit{i.e.}, Open Medical and Healthcare Alliance (Omaha). The data in the graph is highly structured and typically stored in the form of triplets. Since it is difficult for pre-training models to directly learn from triplet information, we sample sub-graphs of different scales and rewrite them into natural language texts.

\begin{wraptable}{l}{6cm}
\centering
\caption{The data statistics involved in general medical knowledge injection.}
\label{pretrain_data}
\begin{tabular}{cc}
\toprule
Data Type        & Token Size \\ \midrule
Medical Books            & 7.18B      \\
Knowledge Graphs & 0.15B      \\
QA Pairs         & 0.24B      \\
Exam Questions   & 0.11B      \\
Articles         & 7.71B      \\ \bottomrule
\end{tabular}
\end{wraptable}

\textbf{Question-Answer Pairs.} We collect publicly available real-world medical question-answer pairs to enrich the training, which is easily accessible and rich medical consultation information in textual form. Therefore, we directly concatenate the question and answer data and use it as training corpora.

\textbf{Exam Questions.} Exam questions usually consist of the question, answer, and explanation. These exam questions are designed to assess medical knowledge and require logical reasoning based on the provided knowledge. Therefore, its data quality is very high, making it an excellent source for pre-training corpora. However, due to specific formatting requirements such as multiple-choice and fill-in-the-blank questions, pre-training models cannot directly consume this data. We also rewrite the exam question data and convert it into medical knowledge points.

\textbf{Articles.} Article data consists of cutting-edge medical papers and popular science articles, which are used to enrich the latest medical knowledge to the dataset. We use PubMed \cite{canese2013pubmed} as the source of the paper data, which is a search engine for medical papers and includes over 25 million articles. PubMed contains both full-text and abstract data. Due to the presence of special text, such as tables in the full-text, which is difficult to process and may interfere with the model's performance, we only use PubMed abstracts as our training data, excluding the abstracts of the expert-level labeled PQA-L (see in Table \ref{pubmedqa_tab}) in PubMedQA dataset for avoiding annotation leakage. For medical popular science articles, we purchase professional medical articles written by different doctors from websites as training corpora.

\textbf{PromptCBLUE.} PromptCBLUE~\citep{zhu2023promptcblue} is a dataset for Chinese medical evaluation jointly constructed by Alibaba Cloud Tianchi, East China Normal University, and other well-known enterprises and academic research institutions, with rankings being set.

\subsubsection{Dataset for Medical Domain Instruction Tuning}
\begin{wraptable}{l}{6cm}
\centering
\caption{The data statistics involved in medical domain instruction tuning.}
\begin{tabular}{cc}
\toprule
Data Type        & Size \\ \midrule
PromptCBLUE            &    68,500   \\ 
Chinese Examination             & 151,568     \\
Wuma QA & 17,604  \\
Huatuo Wiki & 200,000 \\
Multiple-choice Question &  194,455 \\
\hline
\end{tabular}
\end{wraptable}

\begin{table}[!t]
\centering
\caption{Chinese Examination Datasets.}
\label{chineseexam}
\begin{tabular}{cc}
\toprule
\textbf{Chinese Examination}                & \textbf{Size} \\ \midrule
Clinical practitioner                       & 600 \\
Veterinary general practice                 & 400 \\
Licensed pharmacist of Western medicine     & 480 \\ \midrule
\end{tabular}
\end{table}

\textbf{Chinese Examination Datasets.} We collect three examination datasets from the Chinese physician examination data, including clinical practitioner, veterinary general practice, and licensed pharmacist of Western medicine. The overall dataset contains about 150k data. Table \ref{chineseexam} illustrates the distribution of the test set, involving more than 1000 records.

\textbf{Wuma QA.} We additionally collect 17,600 medical question-answer pairs from the realistic environment.

\textbf{Huatuo Wiki.} Huatuo-26M~\citep{li2023huatuo26m} is a large-scale Chinese medical QA dataset. Huatuo Wiki is a subset of Huatuo-26M~\citep{li2023huatuo26m}. The data source of Huatuo Wiki is from the online medical encyclopedia. 

\textbf{Medical Multiple-choice Question Datasets.} Following~\citep{singhal2023expertlevel}, as shown in Table \ref{medqa}, we utilize the MedQA~\citep{jin2021disease}, MedMCQA~\citep{pal2022medmcqa}, PubMedQA~\citep{jin-etal-2019-pubmedqa} and MMLU clinical topics~\citep{hendrycks2020measuring} datasets.

\subsubsection{Dataset for Specific Medical Task Adaptation}
\begin{wraptable}{l}{8.5cm}
\centering
\caption{The detailed information of the PubMedQA dataset.}
\label{pubmedqa_tab}
\begin{tabular}{lccc}
\toprule
\textbf{Statistic}      & \textbf{PQA-L} & \textbf{PQA-U} & \textbf{PQA-A} \\ \midrule
Number of QA pairs      & 1.0k           & 61.2k          & 211.3k         \\
Prop. of yes (\%)       & 55.2           & -              & 92.8           \\
Prop. of no (\%)        & 33.8           & -              & 7.2            \\
Prop. of maybe (\%)     & 11.0           & -              & 0.0            \\ \midrule
Avg. question length    & 14.4           & 15.0           & 16.3           \\
Avg. context length     & 238.9          & 237.3          & 238.0          \\
Avg. long answer length & 43.2           & 45.9           & 41.0          \\ \midrule
\end{tabular}
\end{wraptable}
\textbf{PubMedQA.} As shown in Table \ref{pubmedqa_tab}, the PubMedQA dataset~\citep{jin-etal-2019-pubmedqa} consists of 1,000 question answer pairs annotated by the medical experts and more than 60k unlabeled questions. Given a question with a PubMed abstract as the context, the task is to generate a true choice from the multiple-choice answers (\textit{i.e.}, yes/no/maybe). Besides, PubMedQA requires answer inference from the supporting PubMed abstract context, which is a closed domain question answering task. 




\subsection{Medical Knowledge Modeling}
\label{optimization_tricks}
\begin{figure}[!t]
    \centering
    \includegraphics[width=1\linewidth]{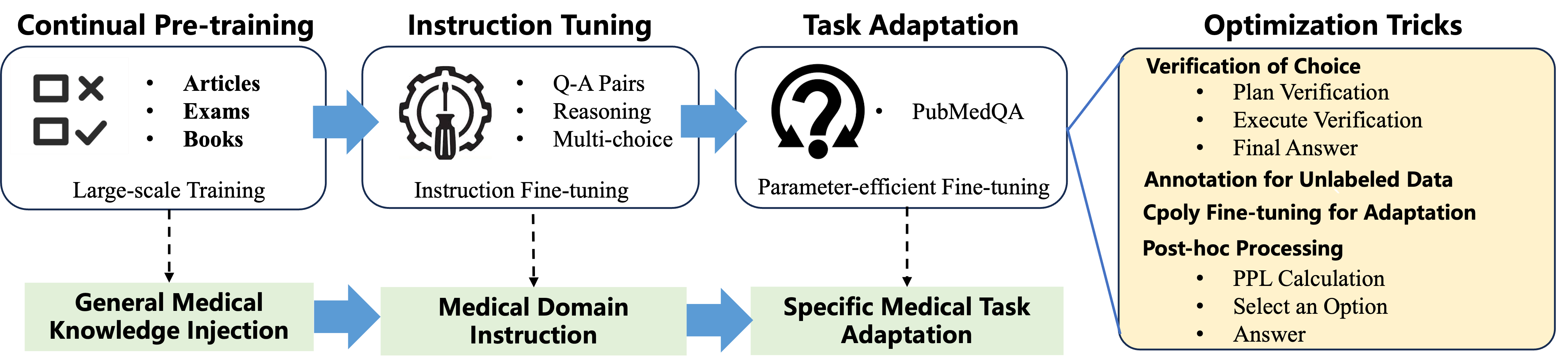}
    \caption{The detailed techniques for different optimization stages.}
    \label{fig:step}
\end{figure}

To take full advantage of medical data, we provide a detailed description for the training tricks for each step of 3-stage optimization. The technique tricks for different steps are shown in Figure \ref{fig:step}.

\subsubsection{Continual Pre-training for Medical Knowledge Injection}
Following \cite{DBLP:conf/acl/DuQLDQY022}, continual pre-training for AntGLM-Med-10B is to optimize an auto-regressive blank-filling objective. Concretely, given an input text $x = $[$x_1$, $x_2$, ..., $x_n$], multiple spans $\{s_1, ..., s_n\}$ are sampled from it as PART B. Then PART B is replaced by the special [MASK] token, resulting in a corrupted sequence $x_{corrupt}$ as PART A. Tokens in Part A can attach each other in a pair-wise way, while tokens in Part B can only attach Part A and the preceding parts in Part B. GLM randomly shuffles the spans to improve the model's semantic understanding for capturing the relationships among different spans. Two-dimensional positional encoding is utilized to represent the positional relationships within each span and among different spans. For auto-regressive generation, each span is padded with special tokens [START] and [END] for input and output, respectively. This approach allows to automatically train both a bidirectional encoder and a unidirectional decoder within a unified architecture. By controlling the span length and granularity, the model can effectively handle both natural language understanding and generation tasks. Due to the limitations of the scale and capability of the general LLM models, we further pre-train the model over the medical data, significantly improving the performance in the medical field.


\textbf{Pre-Training Implementations.} 
The training settings are basically consistent with GLM, using Adam as the optimizer with the following hyper-parameters: $\beta_1$ = 0.9, $\beta_2$ = 0.95. We employ a cosine learning rate schedule, where the final learning rate is equal to 10\% of the maximum learning rate. Additionally, we apply a weight decay of 0.1 and gradient clipping of 1.0. The batch size is set to 256. Since we continue pretraining based on a general base, the learning rate is set to 7e-6. When training a model with a parameter count of 10B, our code achieves a processing speed of approximately 2000 tokens/gpu/sec on 32 A100 GPUs with 80GB RAM.

\begin{figure}[!t]
    \centering
    \includegraphics[width=1\linewidth]{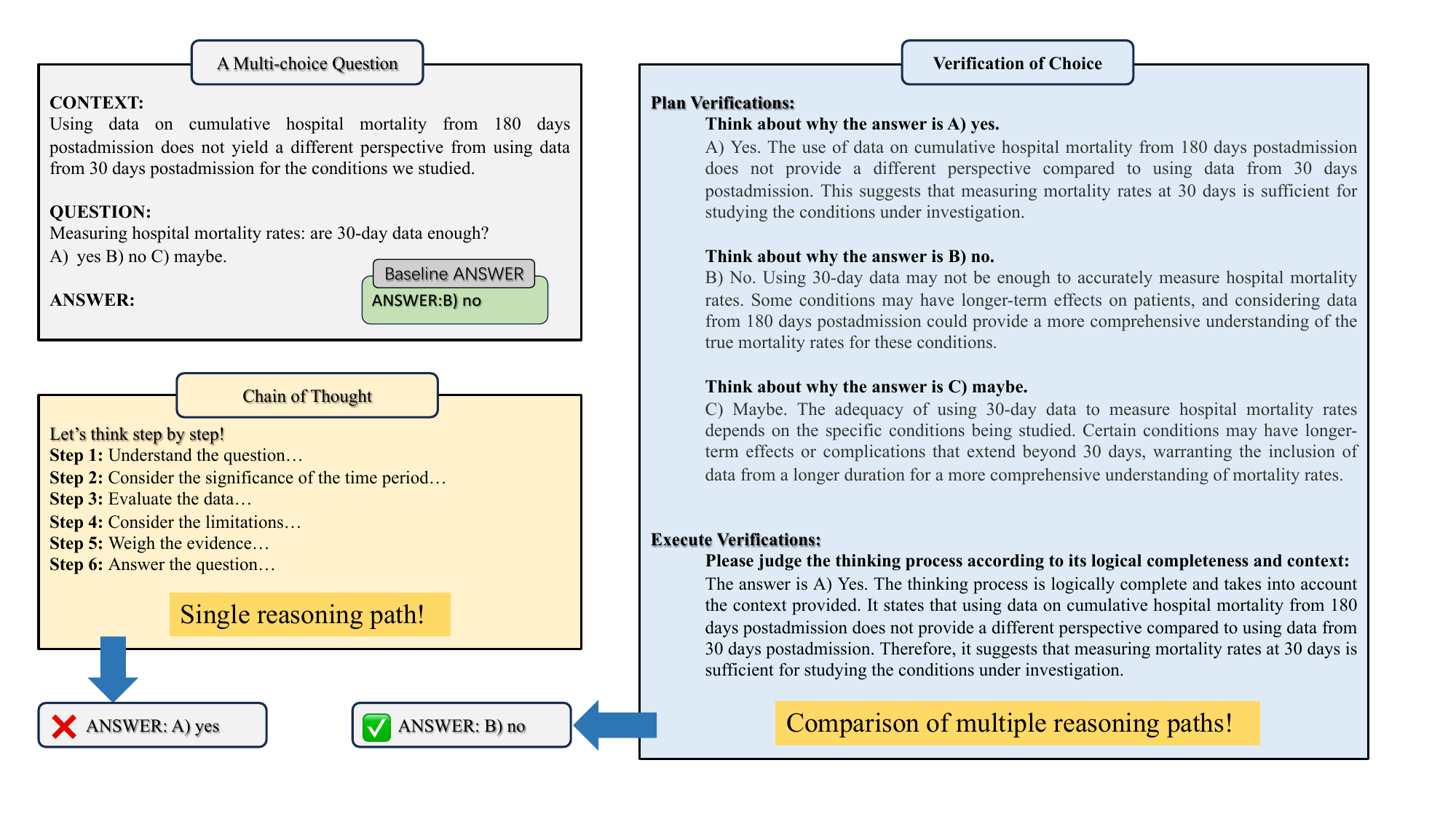}
    \caption{A comparison example for Chain-of-Thought and Verification-of-Choice.}
    \label{fig:VoC}
\end{figure}

\subsubsection{Fine-tuning for Medical Domain Instruction}

\textbf{Instruction Fine-tuning.} We applied instruction fine-tuning for the base LLM following the protocol used by \cite{chung2022scaling}. For this process, we use a cosine learning rate schedule with an initial learning rate of $6e-6$, a weight decay of 0.1 and gradient clipping of 1.0. We utilize 40 A100 GPUs and allocate a batch size of 9 on each device, resulting in an overall batch size of 360.

\textbf{LoRA Fine-tuning.} LoRA, introduced by \citep{hu2022lora}, can decrease the the number of trainable parameters by optimizing pairs of rank-decomposition matrices for pre-trained LLMs, while the original pre-trained model weights are frozen. The approach greatly reduces the storage requirement for LLMs, especially when adapted to some specific tasks. Thus, LoRA enables efficient task-switching without introducing inference latency. For the performance, LoRA can also outperform several other adaptation methods including adapter, prefix-tuning, and fine-tuning. We use the same SFT strategy as described by \citep{hu2022lora} for the PubMedQA benchmark.

\subsubsection{Fine-tuning for Specific Medical Task Adaptation}
In this section, we provide a detailed description for the prompting strategies used for specific medical task adaptation. 

\textbf{Chain-of-Thought} Chain-of-Thought (CoT) is firstly introduced by Wei \textit{et al.} \cite{wei2022chain}, which augments few-shot examples as a enhanced prompt with a step-by-step explanation. The approach empowers an LLM to condition on multi-step outputs towards the final answer. Medical questions usually require a complex multi-step reasoning process, which is well fit for CoT prompting. According to CoT, we exploited a self-generated explanation of each choice for the given medical questions, which provides the reason why an LLM gives this answer.

\textbf{Chain-of-Verification.} Chain-of-Verification (COVE) \cite{dhuliawala2023chain} is proposed to alleviate the hallucination issue, enabling LLMs deliberate on their responses for correcting the mistakes. Given the initial response as a draft, an LLM is required to plan verification questions for fact-checking the draft. Then the LLM should answers those questions independently, guaranteeing each response is not biased by others. Based on the above verification, the LLM can generate the final verified response with a higher confidence.

\textbf{Verification-of-Choice.} Building on COT \cite{wei2022chain} and COVE \cite{dhuliawala2023chain}, we presented a simple prompting strategy named as Verification-of-Choice (VoC) for multi-choice medical questions. VoC involves conditioning an LLM on its own generations for each choice before selecting a choice as the final answer.

The overall process of VoC contains three steps: \textbf{(1)} Plan Multi-Choice Verifications: Given a query, assume that each choice is the true answer, let an LLM self-generate the corresponding explanation for each choice. Each explanation can be taken as a CoT for the choice. \textbf{(2)} Execute Multi-Choice Verifications: Taking multi-choice verifications as a context, LLM could make a comparison among them and self-analyze if there are any inconsistencies. \textbf{(3)} Generate Final Response: Given the inconsistencies (if any) among multi-choice verification, LLM could generate a final response.

Unlike CoT and COVE, VoC may be used to investigate the inconsistency between the explanations of multiple choices and questions, thus it is helpful to produce more accurate responses. For example, the explanation of a choice is different with the conditions in the question. In this work, we apply VoC only for multiple-choice question task.

\textbf{Post-hoc Processing.}
PPL ranking is adopted as the indicator of uncertainty. For multi-choice QA datasets, the specific solution is to concatenate the original text with each option. Then a LLM is required to calculate the corresponding perplexity (Perplexity score) and select the option with the smallest PPL as the final predicted option. As for other datasets, generation and post-processing are utilized to produce the final response.

\textbf{LLM-annotated PQA-U.} As shown in Table \ref{pubmedqa_tab}, the PQA-U subset of the PubMedQA dataset is not well utilized due to without accurate annotations. It is noticed that long answers of the PQA-U subset are provided, which implicitly involve the correct answer for each question. To fully utilize the PQA-U subset, given a question in PQA-U, LLMs are required to generate responses for questions with the corresponding long answer, where VoC strategy is used to improve the correctness of answers. Then the generate answers are taken as pseudo annotations of PQA-U, which join the stage of specific medical task adaptation. 

To evaluate the annotation accuracy for PQA-U, we fine-tuned two models and reported their performance on the test set of PubMedQA. Concretely, one model is fine-tuned on the PQA-L and PQA-U annotated using long answers. Another model is fine-tuned on the PQA-L and PQA-U annotated using both long answers and VoC. The results are reported in Section.

\begin{figure}[!t]
\centering
\begin{minipage}[t]{0.48\textwidth}
\centering
\includegraphics[width=7cm]{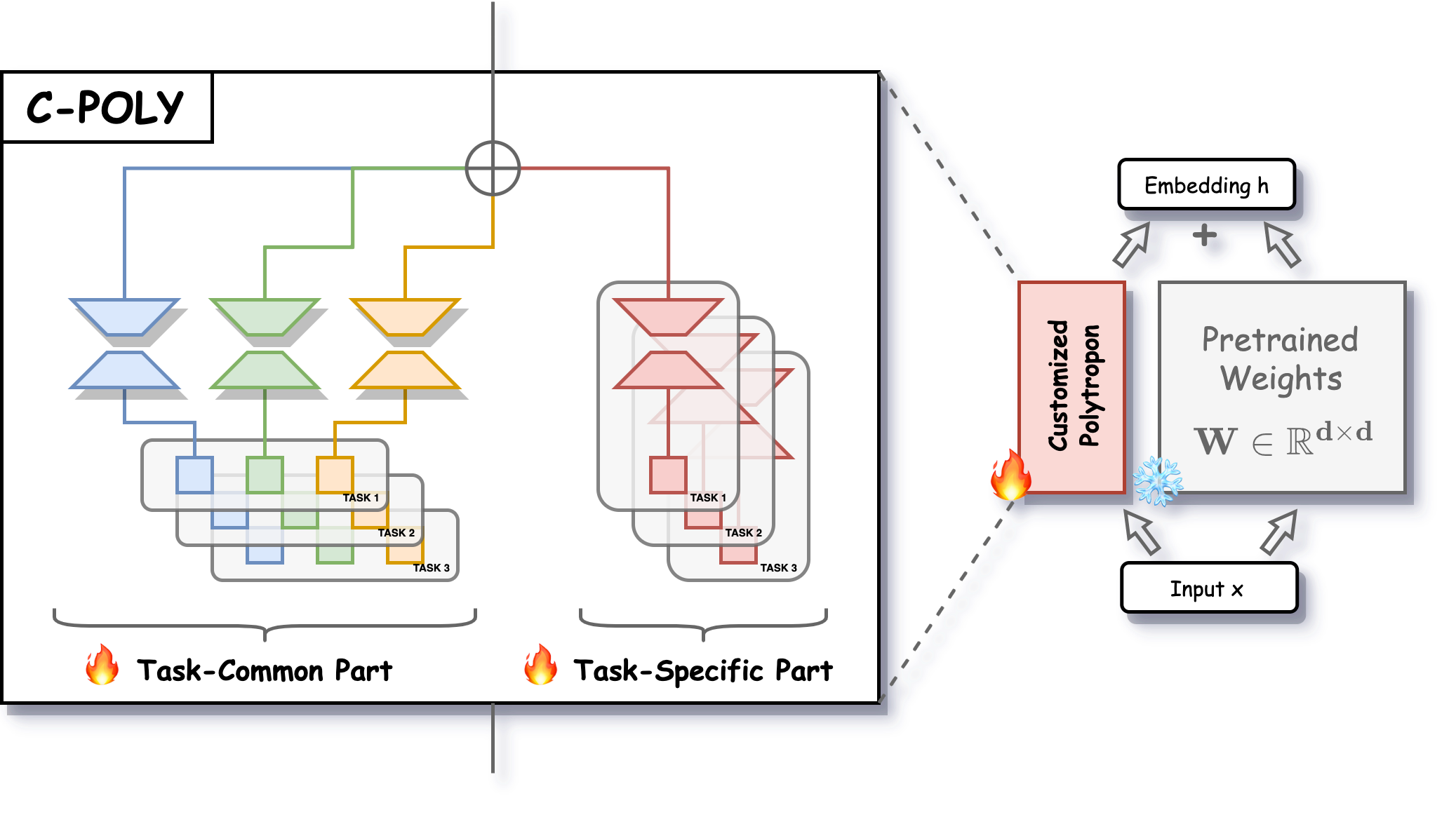}
\caption{Overview of $\texttt{C-Poly}$ framework.}
\label{fig:cpoly_architecture}
\end{minipage}
\begin{minipage}[t]{0.48\textwidth}
\centering
\includegraphics[width=5.5cm]{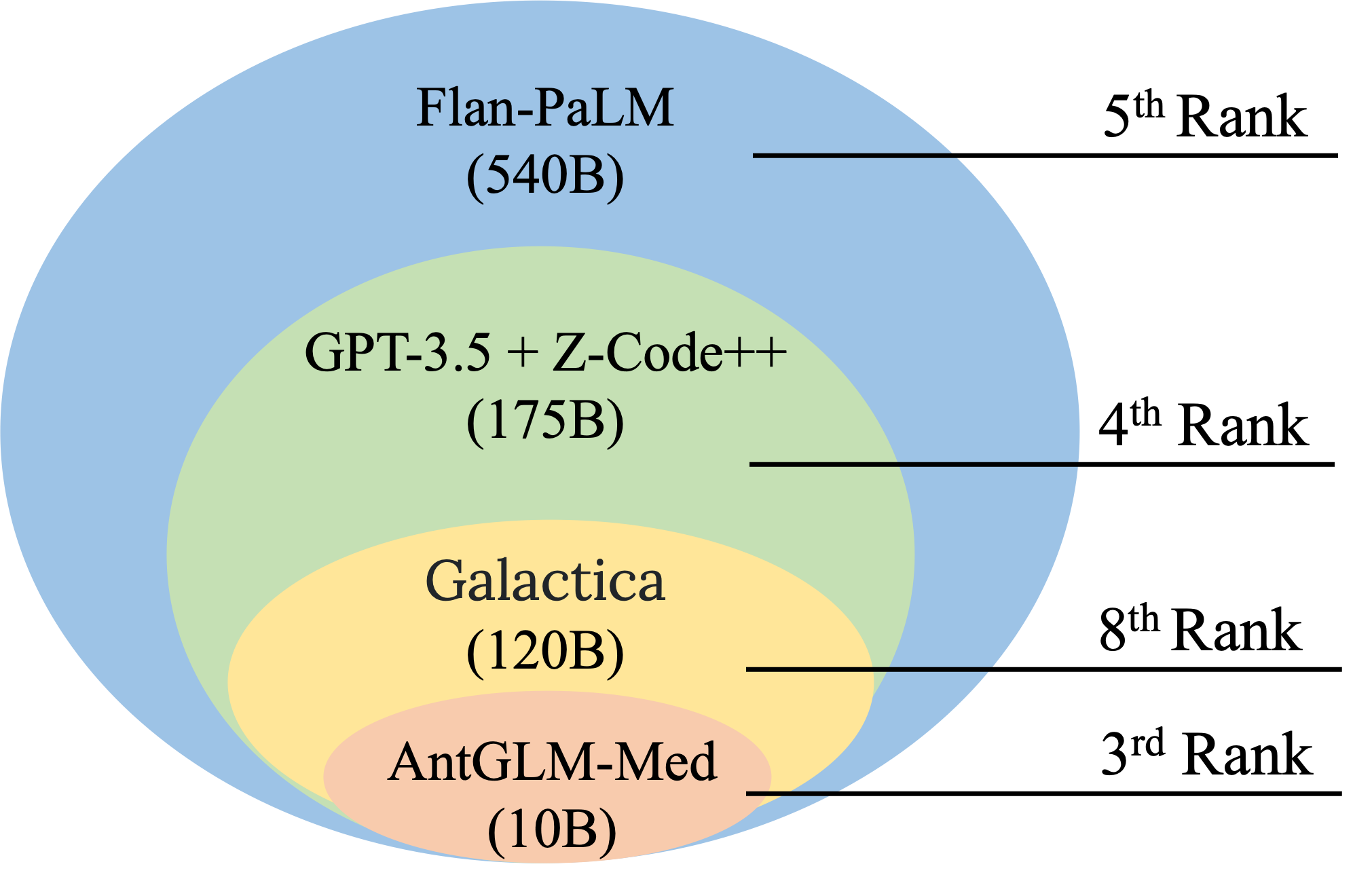}
\caption{The PubMedQA learderboard.}
\label{lead_fig}
\end{minipage}
\end{figure}

\textbf{CPoly Fine-tuning.} $\texttt{C-Poly}$, a fine-tuning method by \cite{iclr2023customizable}, uses multi-task learning and adapters to differentiate shared and customized skills. It allows multi-class task samples to learn from each other in multi-adapter PEFT. The unified MTL framework $\texttt{C-Poly}$ (Figure~\ref{fig:cpoly_architecture}) enhances sample efficiency across tasks. For $T$ tasks with data $x^t$, the PLE-like structure has adapter modules $\mPhi$ with $|\mPhi_A| + T \times |\mPhi_B^t| = A + T \times B$ adapters, assuming $B=1$. The $\texttt{C-Poly}$ output is the sum of shared and task-specific adapters (Equation~\ref{eq:cpoly}), with $w_i$ as learnable weights.
\begin{equation}
\label{eq:cpoly}
        \underbrace{\sum_{i=1}^{A}{w^{t}_{i}\phi_{i}(x^t)}}_{\text{Task-Common}} + \underbrace{\sum_{j=1}^{B}w^{t}_{j}\phi^{t}_{j}(x^t)}_{\text{Task-Specific}}
        = \sum_{i=1}^{A}{w^{t}_{i}\phi_{i}(x^t)} + 
        w^{t}\phi^{t}(x^t)
\end{equation}

The allocation matrix $\mW \in \R^{T \times (A + T)}$ differentiates shared ($\mW_A$) and customized ($\mW_B$) skills. Different learning methods optimize skill acquisition, using low-rank approximations for efficiency. Shared skills use a Gumbel-sigmoid approach for differentiable sampling. Specialized skills are learned by differentiating shared and exclusive modules.

By combining the above-mentioned three optimization stages together, we fine-tine AntGLM-10B on the collected large-scale, high-quality, medical-domain corpus, resulting in our final model AntGLM-Med-10B.


\begin{table}[!t]
\centering
\label{lead_pubmedqa}
\caption{The leaderboard for the PubMedQA dataset. Our AntGLM-Med-10B could obtain competitive performance with a relatively small model size.}
\begin{tabular}{lcc}
\toprule
Model                                                                            & Size & Accuracy \\ \midrule
Med-PaLM 2~\citep{singhal2023expertlevel}                                         & NA   & 81.8     \\
Palmyra-Med~\citep{kamble2023palmyra}                                                  & 40B  & 81.1     \\
\color{red} AntGLM-Med                                                                               & \color{red}10B   & \color{red}80.6 \\
GPT-4-base~\citep{nori2023capabilities}                                                & NA   & 80.4     \\
GPT-3.5 + Z-Code++~\citep{he2023zcode}                                        & 175B & 79.6     \\
Flan-PaLM-3-shot~\citep{singhal2022large}                                   & 540B & 79.0     \\
Codex-5-shot~\citep{liévin2023large} & 175B & 78.2     \\
Human Performance~\citep{jin-etal-2019-pubmedqa}       & NA   & 78.0     \\
Galactica~\citep{taylor2022galactica}                                                           & 120B & 77.6     \\
GatorTronGPT~\citep{peng2023study}                                   & 20B  & 77.6    \\ \bottomrule
\end{tabular}
\end{table}

\section{Experiments}
\label{experiment}
\subsection{Experimental Setup}
In our experiments, we mainly discuss the LLM's performance after the final stage (Specific Medical Task Adaptation). The details are described as follows. In the case of vanilla LoRA, we set the rank of the low-rank approximation, $r=8$. We utilize $4$ parallel LoRA for task-common skills and $1$ LoRA for task-custom skill in $\texttt{CPoly}$, the rank is set to $r=4$. This decision is made to ensure a comparable number of training parameters across all methods. We train our model for $10$ epoch with a batch size of $12$ on given datasets during training. The AdamW optimizer~\citep{loshchilov2017decoupled} is used with a learning rate of $5e^{-5}$. We also employ the linear decay strategy~\citep{loshchilov2016sgdr} as the learning rate scheduler with a weight decay of $0.01$ and a warming up ratio of $0.06$. 


\subsection{Model Evaluation}
To evaluate our AntGLM-Med-10B model, we consider multi-choice question as the adaptation task in the third optimization stage). The PubMedQA dataset is utilized as the evaluation dataset, which requires medical research comprehension skills to make reasoning over PubMed abstract context. We use 500 test samples for evaluation.

\subsection{Experimental Results}
\subsubsection{Main Results}
\textbf{Results for PubMedQA.} On PubMedQA, AntGLM-Med-10B obtained a score of 80.6\%. This is below the state-of-the-art performance (81.8 from Med-PaLM 2~\citep{singhal2023expertlevel}) and second place (81.1 from Palmyra-Med~\citep{kamble2023palmyra}). The main reason is due to the relatively smaller model size, \textit{e.g.}, AntGLM-Med-10B \textit{vs.} Palmyra-Med-40B. Although that, AntGLM-Med-10B could exhibit improved performance compared to some other larger LLMs (as shown in Figure \ref{lead_fig}). Besides, AntGLM-Med-10B surpasses the models with similar model sizes (\textit{e.g.}, GatorTronGPT~\citep{peng2023study}).

\begin{figure}[!t]
    \centering
    \includegraphics[width=0.65\linewidth]{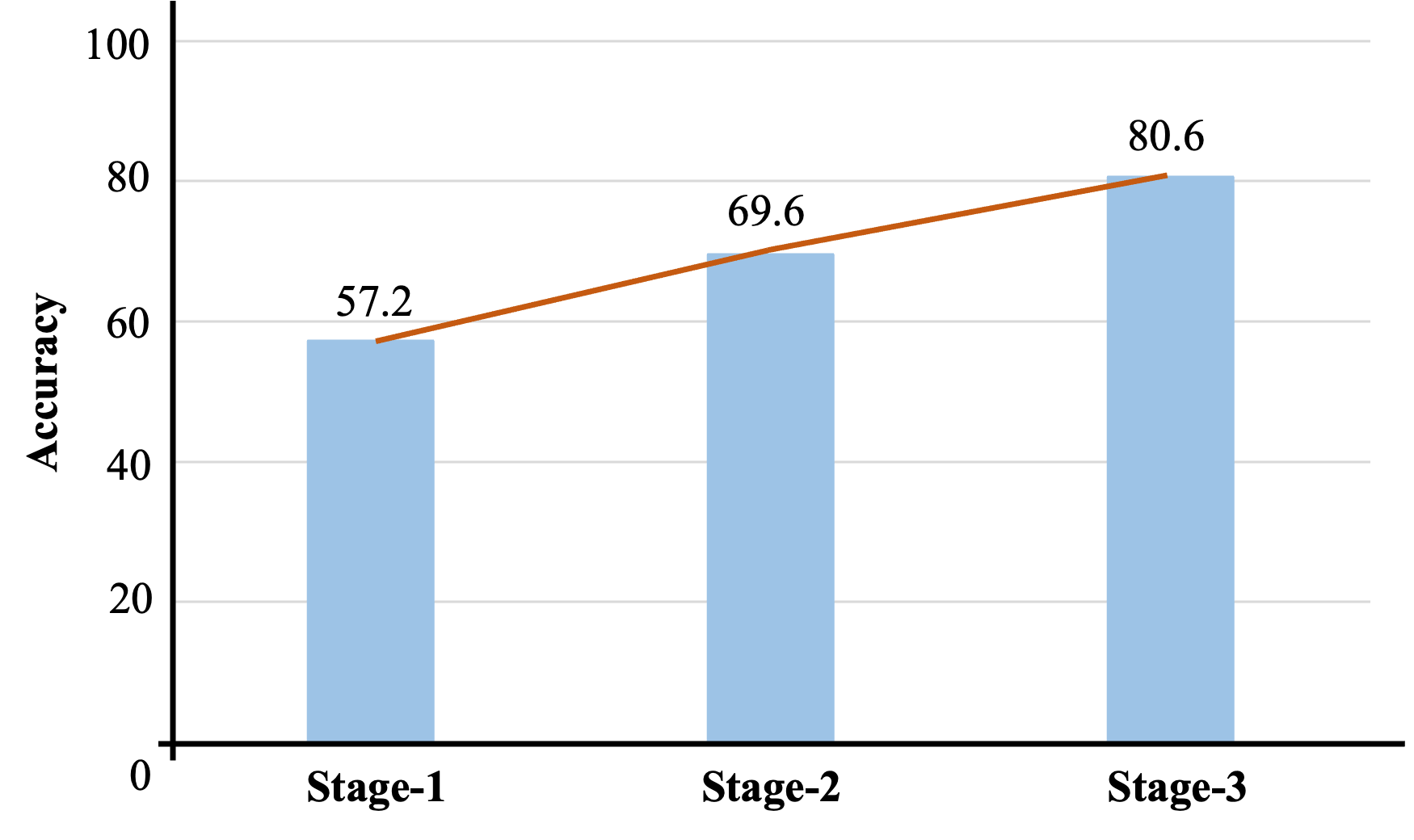}
    \caption{Accuracy results on PubMedQA at different optimization stages. Stage-1 is the accuracy based on general medical knowledge injection. Stage-2 is for medical domain instruction based on Stage-1. Stage-3 is for specific medical task adaptation based on Stage-2.}
    \label{acc:step}
\end{figure}

\textbf{Results for Different Stages.} As shown in Figure \ref{acc:step}, we observed continuous accuracy improvements on the PubMedQA dataset throughout the entire optimization procedure. It is observed that the base model only obtained an initial accuracy of 57.2\% on PubMedQA, and ultimately achieved a 80.6\% accuracy by the end of the optimization procedure, which indicate the effectiveness of the 3-stage optimization.

\begin{table}[!t]
\centering
\caption{The accuracy results for models fine-tuned on PQA-L and PQA-U. PQA-U are annotated using different strategies. Long answer and VoC can provided the best accuracy results for LoRA and full parameter fine-tuning adaptation strategies, indicating that VoC can further improve the annotation accuracy for PQA-U.}
\label{tab:my-voc-ablation-table}
\begin{tabular}{@{}clc@{}}
\toprule
\textbf{Fine-tuning}  & \textbf{Annotation Strategy}  & \textbf{PubMedQA} \\ \midrule
\multirow{2}{*}{LoRA} & Long Answer          & 86.6 \\ \cmidrule{2-3}
                      & Long Answer + VoC    & 87.6 \\ \midrule
\multirow{2}{*}{Full Parameter Tuning}  & Long Answer          & 87.8    \\ \cmidrule{2-3}
                       & Long Answer + VoC     & 88.2  \\ \bottomrule
\end{tabular}
\end{table}

\begin{table}[!t]
\centering
\caption{The detailed performance for specific task adaptation using different datasets and tuning strategies.}
\label{tab:my-dataset-ablation-table}
\begin{tabular}{@{}clcc@{}}
\toprule
\textbf{Fine-tuning} & \textbf{Dataset} & \textbf{Size} & \textbf{PubMedQA} \\ \midrule
\multirow{9}{*}{Cpoly} & PubMedQA & 500 & \multirow{3}{*}{80.6} \\ 
                       & MedMCQA  & 500 &                       \\
                       & MedQA    & 500 &                       \\ \cmidrule{2-4}
                       & PubMedQA & 500 & \multirow{2}{*}{80.4} \\ 
                       & MedMCQA  & 500 &                       \\ \cmidrule{2-4}
                       & PubMedQA & 500 & \multirow{2}{*}{80.2} \\ 
                       & MedQA    & 500 &                       \\ \cmidrule{2-4}
                       & MedMCQA  & 500 & \multirow{2}{*}{-}    \\
                       & MedQA    & 500 &                       \\ \midrule
\multirow{3}{*}{LoRA}  & PubMedQA & 500 & 78.8  \\ \cmidrule{2-4}
                       & MedMCQA  & 500 & 78.6  \\ \cmidrule{2-4}
                       & MedQA    & 500 & 78.0  \\ \bottomrule
\end{tabular}
\end{table}

\subsubsection{More Discussions}
\textbf{Effectiveness of VOC.} \label{ablation_voc} To evaluate the annotation accuracy for PQA-U, we fine-tuned two LLM models and reported their performance on the test set of PubMedQA. Concretely, one model is fine-tuned on the PQA-L and PQA-U, where PQA-U is annotated only using long answers. Another model is fine-tuned on the PQA-L and PQA-U, where PQA-U is annotated using both long answers and VoC. The results are reported in Table \ref{tab:my-voc-ablation-table}. Using both long answer and VoC to annotate PQA-U, the model can exhibit higher accuracy, which indicate that VoC can help to improve the annotation accuracy for PQA-U.

\textbf{Effectiveness of Cpoly.} We evaluated the effectiveness of $\texttt{CPoly}$ on multiple datasets and conducted an ablation experiment in Table~\ref{tab:my-dataset-ablation-table}. We randomly selected 500 samples from the PubMedQA, MedMCQA, and MedQA datasets respectively, and combined them using different strategies. We then validated the training results of a single, two, and three dataset samples, respectively. Since the router vectors of multiple adapters trained in $\texttt{CPoly}$ cannot index and effectively predict untrained unknown tasks, we only reported the dataset's performance with the corresponding training set when using CPoly. The training results of $\texttt{CPoly}$ on multiple datasets achieved significantly more improvement than those of LoRA on a single dataset. As the number of datasets involved in the $\texttt{CPoly}$ training process increases, the best performance on PubMedQA increases. As the number of datasets trained with multitasking increased, the performance is significantly improved compared with training on a single dataset.

\section{Conclusion}
In this work, we explore how to adapt a pre-trained general LLM in medical domain, from a medical beginner to a medical expert. Starting from a pre-trained general LLM model (AntGLM-10B), we leverage a 3-stage optimization procedure to fine-tune it, \textit{i.e.}, continual pre-training for medical knowledge injection, medical domain instruction tuning, and specific medical task adaptation. Different large-scale medical datasets are collected, covering various data types and different tasks, such as question-answering (QA), medical reasoning, multi-choice question, and medical conversations. Besides, for multi-choice question in medical domain, we design a novel Verification-of-Choice approach for prompting engineering, which can significantly enhance the reasoning ability of LLMs. By combining the above points together, our AntGLM-Med-10B model can exhibit competitive performance compared with both general LLMs and other LLMs pre-trained on medical knowledge on the PubMedQA. It is noticed that AntGLM-Med-10B can outperform the most of larger LLMs (>40B).

\begin{table}[!th]
\centering
\label{promp_pubmedqa}
\caption{A prompt example for generating LLM-annotated PQA-U on PubMedQA dataset.}
\begin{tabular}{cl}
\toprule
\textbf{Context}& 
\begin{tabular}[c]{p{0.7\linewidth} p{0.05\linewidth}}
\textbf{OBJECTIVE.} We compare 30-day and 180-day postadmission hospital mortality rates for all Medicare patients and those in three categories of cardiac care: coronary artery bypass graft surgery, acute myocardial infarction, and congestive heart failure.
\\
\textbf{DATA SOURCES/COLLECTION.} Health Care Financing Administration (HCFA) hospital mortality data for FY 1989.
\\
\textbf{STUDY DESIGN.} Using hospital level public use files of actual and predicted mortality at 30 and 180 days, we constructed residual mortality measures for each hospital. We ranked hospitals and used receiver operating characteristic (ROC) curves to compare 0-30, 31-180, and 0-180-day postadmission mortality.
\\
\textbf{PRINCIPAL FINDINGS.} For the admissions we studied, we found a broad range of hospital performance when we ranked hospitals using the 30-day data; some hospitals had much lower than predicted 30-day mortality rates, while others had much higher than predicted mortality rates. Data from the time period 31-180 days postadmission yield results that corroborate the 0-30 day postadmission data. Moreover, we found evidence that hospital performance on one condition is related to performance on the other conditions, but that the correlation is much weaker in the 31-180-day interval than in the 0-30-day period. Using ROC curves, we found that the 30-day data discriminated the top and bottom fifths of the 180-day data extremely well, especially for AMI outcomes.
\\
\textbf{QUESTION:} Measuring hospital mortality rates: are 30-day enough?
\\
A) yes  B) no  C) maybe
\end{tabular}
\\ \midrule

\textbf{Verification-of-Choice} & 
\begin{tabular}[c]{p{0.7\linewidth} p{0.05\linewidth}}
\textbf{Step-1: Plan Multi-Choice Verification}
\\
Think about why the answer is A) yes.

A) Yes. The use of data on cumulative hospital mortality from 180 days postadmission does not provide a different perspective compared to using data from 30 days postadmission. This suggests that measuring mortality rates at 30 days is sufficient for studying the conditions under investigation.

Think about why the answer is B) no.

B) No. Using 30-day data may not be enough to accurately measure hospital mortality rates. Some conditions may have longer-term effects on patients, and considering data from 180 days postadmission could provide a more comprehensive understanding of the true mortality rates for these conditions.

Think about why the answer is C) maybe.

C) Maybe. The adequacy of using 30-day data to measure hospital mortality rates depends on the specific conditions being studied. Certain conditions may have longer-term effects or complications that extend beyond 30 days, warranting the inclusion of data from a longer duration for a more comprehensive understanding of mortality rates
\\ \midrule
\textbf{Step-2: Execute Multi-Choice Verification}

Please judge the yes/no/maybe thinking process according to its logical completeness and context. 

The answer is A) Yes. The thinking process is logically complete and takes into account the context provided. It states that using data on cumulative hospital mortality from 180 days postadmission does not provide a different perspective compared to using data from 30 days postadmission. Therefore, it suggests that measuring mortality rates at 30 days is sufficient for studying the conditions under investigation.
\\ \midrule

\textbf{Step-3: Generate Final Response}\\
{\color{green}\textbf{Answer: A}} 
\end{tabular}
\\ \bottomrule
\end{tabular}
\end{table}

\newpage

\bibliography{iclr2022_conference}
\bibliographystyle{iclr2024_conference}

\end{document}

%% file: math_commands.tex
\usepackage{amsmath,amsfonts,bm}

\def\eqref#1{equation~\ref{#1}}

\def\1{\bm{1}}

\def\mW{{\bm{W}}}

\def\mPhi{{\bm{\Phi}}}

\DeclareMathAlphabet{\mathsfit}{\encodingdefault}{\sfdefault}{m}{sl}
\SetMathAlphabet{\mathsfit}{bold}{\encodingdefault}{\sfdefault}{bx}{n}

\newcommand{\R}{\mathbb{R}}



%% file: packages.tex
\usepackage{hyperref}
\usepackage{url}

\usepackage{graphicx}
\usepackage{}
\usepackage{booktabs}
\usepackage[symbol]{footmisc}
\renewcommand{\thefootnote}{\fnsymbol{footnote}}
\usepackage{amssymb}
\usepackage{pifont}
\usepackage{graphicx}  
\usepackage{amsmath} 
\usepackage{amssymb}
\usepackage{mathabx}
\usepackage{hyperref}
\usepackage{glossaries}
\usepackage{enumitem}
\usepackage{arydshln}
\usepackage{makecell}
\usepackage{longtable, tabu}
\usepackage{enumitem}
\usepackage{placeins}
\usepackage{array}
\usepackage{multirow}
\usepackage[T1]{fontenc}
\usepackage[font=small,labelfont=bf,tableposition=top]{caption}
\usepackage{etoolbox}
\usepackage[utf8]{inputenc}
\usepackage{float}
\usepackage{wrapfig,lipsum,booktabs}
\usepackage{caption}
\usepackage{enumitem}
\usepackage{blindtext}
\usepackage{array}
\usepackage{bbm}
\usepackage{todonotes}
\usepackage{cleveref}

\renewcommand{\thefootnote}{\arabic{footnote}}

\usepackage{xcolor}
\usepackage{soul}
\usepackage{multicol}
\usepackage{tabularx,colortbl}
\usepackage{bbm}

\newcolumntype{L}{>{\ttfamily}l}

\definecolor{COLORA}{rgb}{0.33999999999999997, 0.86, 0.3712}
\definecolor{COLORB}{rgb}{0.33999999999999997, 0.8287999999999999, 0.86}
\definecolor{COLORC}{rgb}{0.8287999999999999, 0.86, 0.33999999999999997}
\definecolor{COLORD}{rgb}{0.86, 0.33999999999999997, 0.8287999999999999}
\definecolor{COLORE}{rgb}{0.3712, 0.33999999999999997, 0.86}
\definecolor{COLORF}{rgb}{0.86, 0.3712, 0.33999999999999997}
\definecolor{COLORG}{rgb}{0.5, 0.5, 0.5}

\definecolor{niceblueshade}{HTML}{0C6DC7}   
\definecolor{niceblue}{HTML}{1F5B93}   
\definecolor{nicered}{HTML}{BE533B}  
\definecolor{nicegreen}{HTML}{54AD72}  
\definecolor{nicegray}{rgb}{0.3, 0.3, 0.3}
